\documentclass[linguex]{sp}

\usepackage{graphicx}  
\usepackage{times}
\usepackage{natbib}
\usepackage{amsmath}
\usepackage{amssymb}
\usepackage{amsthm}
\usepackage{xspace}
\usepackage{soul}
\usepackage{booktabs}
\usepackage{esvect}
\usepackage{stmaryrd}
\usepackage{MnSymbol}
\usepackage{linguex}
\usepackage{drs}
\usepackage{hyperref}
\hypersetup{
    colorlinks,
    citecolor=black,
    filecolor=black,
    linkcolor=black,
    urlcolor=black
  }
\usepackage{listings}

\usepackage{mydefs}

\title{How to marry a star:\\ probabilistic constraints for meaning in context}

\author[Katrin Erk \& Aur\'elie Herbelot]{
  \spauthor{Katrin Erk\\  \institute{University of Texas at Austin}} \AND 
  \spauthor{Aur\'elie Herbelot\\ \institute{University of Trento}}}

\begin{document}
\maketitle

\begin{abstract}
  In this paper, we derive a notion of \textit{word meaning in context} that characterizes meaning as both intensional and conceptual.   We introduce a framework for specifying local as well as global constraints on  word meaning in context, together with their interactions, thus modelling the wide range of lexical shifts and ambiguities observed in utterance interpretation.  We represent sentence meaning as a \textit{situation description system}, a probabilistic  model which takes utterance understanding to be the mental process of describing to oneself one or more situations that would account for an observed utterance. We show how the system can be implemented in practice, and apply it to examples containing various contextualisation phenomena.
\end{abstract}

\section{Introduction}
\label{sec:introduction}

Word meaning is flexible. This flexibility is often characterised by distinguishing the `context-independent' meaning of a lexical item (its definition(s) in a dictionary) and its `speech act' or `token' meaning -- the one it acquires by virtue of being used in the context of a particular sentence \citep{Grice1968}. The generation of a token meaning goes well beyond word sense disambiguation and typically involves speakers' knowledge of the world as well as their linguistic knowledge. For instance, \citet[pp.222-223]{Searle1980} reminds us that \textit{to cut grass} and \textit{to cut a cake} evoke different tools in the mind of the comprehender (a lawnmower vs a knife). 

The question of context dependence is associated with long-standing debates in both linguistics and philosophy, with theoretical positions ranging from semantic minimalism to radical contextualism. Our goal in this paper is not to take a side in those debates, but rather to give an integrated account of the many different ways context interacts with lexical meaning. In particular, we will set up formal tools to talk about the dependencies that exist between the lexicon and the various layers involved in utterance interpretation, from logical effects to situational knowledge.


Let us first consider what contextual influences might play a role in shifting the meaning of a word. The first effect that comes to mind might be \textit{local context}. Specific combinations of predicates and arguments activate given senses of the lexical items involved in the composition. This is known as `selectional preference' and can be demonstrated with the following example:

\ex. \label{ex:blade} She drew a blade.

In this case, where words in both the predicate and the argument positions have multiple senses, the sentence can mean that the agent sketched either a weapon or a piece of grass, or that she randomly sampled either a weapon or a piece of grass, or that she pulled a weapon out of a sheath (but probably not a piece of grass). In this example, both the predicate and argument are ambiguous, and they seem to restrict each other's senses, which then makes the ``pull out a piece of grass'' reading unavailable. 

But word meaning is not only influenced by semantic-role neighbors. \textit{Global context} is involved. \ref{ex:ray} is a contrast pair adapted from an example by Ray Mooney (p.c.), with different senses of the word \textit{ball} (sports equipment vs dancing event). Arguably, the sense of the predicate \textit{run} is the same in \ref{ex:ray_a} and \ref{ex:ray_b}, so the difference in the senses of \textit{ball} must come from something other than the syntactic neighbors, some global topical context brought about by the presence of \textit{athlete} in the first sentence, and \textit{violinist} in the second. 

\ex. \label{ex:ray} \a. \label{ex:ray_a} The athlete ran to the ball.
\b. \label{ex:ray_b} The violinist ran to the ball. 

There is even a whole genre of jokes resting on a \textit{competition of local and global topical constraints} on meaning: the pun. Sentence \ref{ex:astronomer} shows an example.

\ex. \label{ex:astronomer} The astronomer married the star. 

This pun rests on two senses of the word \textit{star}, which can be paraphrased as `well-known person' and `sun'. It is interesting that this sentence should even work as a pun: The predicate that applies to \textit{star}, \textit{marry}, clearly selects for a person as its theme. So if the influence of local context were to apply strictly before global context, \textit{marry} should immediately disambiguate \textit{star} towards the `person' sense as soon as they combine. But the `sun' sense is clearly present.\footnote{In fact, our own intuitions about sentence \ref{ex:astronomer} vary. One of us prominently perceives the reading where the astronomer weds a gigantic ball of fire; for the other one of us, the sentence oscillates between the two different senses of \textit{star}.} In other words, local context and global topical context seem to be competing.

If lexical meaning cannot easily be pinned down to sense disambiguation, and if it is indeed dependent on the interaction of a number of constraints that may go beyond the lexicon, a model of meaning in context should answer at least two core questions: Is it possible to predict not one, but all the interpretations a set of speakers might attribute to a word or phrase? How does the interaction of various constraints take place in the shift from context-independent to token meaning? This paper takes on the task of formalising a semantic framework which accounts for the wide flexibility of word meaning and the range of interpretations it can take in a given sentence. We frame the question as modelling the comprehension process undergone by the hearer of an utterance: we ask what kind of `meaning hypotheses' are invoked by a listener when presented with a given utterance, in particular, what those hypotheses are made of, and how they relate to sentence constituents. 

In our framework, which we call  \textit{situation description systems}, we draw on two main inspirations. On the theoretical side, we draw on work in linguistics and cognition that takes lexical knowledge and knowledge about wider situations to be inextricably linked. Notable advocates of such accounts include \citet{Fillmore:Usemantics}, who develops a conception of sentence understanding where the words in a sentence evoke chunks of background knowledge, which the listener integrates into a single whole. \citet{SanfordGarrod}, and more recently \citet{McRae:LLC}, also argue that knowledge of scenarios is used during sentence processing. Likewise, we will describe sentence understanding involving mental concepts as well as scenarios.


For the formalization, we draw on probabilistic graphical models. Such models describe complex joint probability distributions as a graph where edges indicate dependencies, or probabilistic constraints. They are designed to handle constellations with multiple sources of uncertainty that mutually constrain each other, and are therefore well suited to represent a distribution over outcomes rather than a single label. In the context of this paper, we propose to use them to probabilistically infer different competing interpretations for a word in context. This will allow us to model word meaning as a distribution over hypotheses, constrained by the interplay between sentence meaning on the one hand, and the concepts underlying the words in the sentence on the other hand.

The focus of the present paper is on formalization. We will illustrate the behaviour of our system with simple utterances containing verbs and object-denoting nouns, inspecting whether our system implementation outputs results that match our intuition. In particular, we want to observe that multiple senses can be activated during comprehension, for instance when processing a pun. For more commonplace sentences such as \textit{a player was carrying a bat}, we would like more of the probability mass going to the sport equipment interpretation of \textit{bat}, while reserving some readings for the unlikely case where the player carries an actual animal. We aim to expand the system in the future, covering more complex sentences and longer stretches of discourse. Ultimately, we envisage that the full system could be evaluated with respect to its ability to simulate human behaviour on tasks like sense annotation or lexical substitution. 

In what follows, we first give a brief introduction to probabilistic graphical models and highlight how they can be used in linguistic settings (\S\ref{sec:generative}). We then proceed with a general formalization of situation description systems (\S\ref{sec:situationdescriptions}). \S\ref{sec:sds_selpref} through \S\ref{sec:argument} form the core of the paper where we discuss examples and formulate specific probabilistic constraints to account for lexical meaning. Finally we speculate about extensions in \S\ref{sec:extensions}. 

\section{Probabilistic graphical models for situation descriptions}
\label{sec:generative}

Let us consider the following sentences:

\ex.\label{ex:illustrations} 
\a. \label{ex:simple} A bat was sleeping.
\b. \label{ex:ambiguous} A player was holding a bat.
\c. \label{ex:situation} A girl was eating.
\d. \label{ex:astronomer_illu} An astronomer married a star.
\e. \label{ex:argument} She seems to revel in  arguments  and loses no opportunity to
declare her political principles.\footnote{This sentence is from the word sense disambiguation dataset of \citet{SE3}.}

Examples \ref{ex:simple}-\ref{ex:ambiguous}, as well as \ref{ex:argument}, involve different senses of the words \textit{bat} and \textit{argument}, at different levels of granularity. \ref{ex:simple} is a straightforward case of selectional constraint, where the agent of \textit{sleep} is assumed to be an animate being. \ref{ex:ambiguous}, on the other hand, cannot be directly resolved by selectional constraint, but rather by looking at the wider context of the sentence: the sense of \textit{bat}, the patient of \textit{hold}, is dependent on the agent of the same verb. Interestingly, selectional constraint and wider context can be at odds with each other, and this is illustrated in \ref{ex:astronomer_illu}, where the sense of \textit{star} oscillates, as discussed in our introduction. In \ref{ex:argument}, the precise interpretation of \textit{argument} is dependent on the meanings attributed to the other lexical items in the sentence, but also wider context, such as the connotation of the phrase \textit{political principles} in the second half of the utterance. Finally,  \ref{ex:situation}, whilst being unambiguous, suggests specific entailments with higher probability than others (the girl is more likely to be eating an apple than a stone). 

What is common to all those cases? They illustrate how a word's sense is influenced by \textit{some} aspect(s) of the linguistic structure of the entire sentence. They also show that in many cases, the sense of the word, as well as the specific entailments it affords, are uncertain and could take different values. Even in a seemingly straightforward example such as \ref{ex:ambiguous}, there is some small chance that the player is actually holding an animal rather than a baseball bat, and given the right discourse context (a Harry Potter novel, for example), that sense could be activated. That is, we want to be able to express meanings probabilistically, with respect to the variety of interactions that can take place within a sentence. A natural tool to achieve this is the \textit{probabilistic graphical model}, as we will now see.

\subsection{A short introduction to probabilistic graphical models}

Probabilistic graphical models form the technical core of our approach. They let us represent concepts underlying content words as nodes in a graph, where edges represent weighted, interacting constraints. Depending on the specific way the graph is navigated, particular topics, world knowledge and lexical content are activated and thus give rise to the individual meanings of words, as well as the overall interpretation of the utterance under consideration.

\begin{figure}[tb]
  \centering
  \includegraphics[scale=0.4]{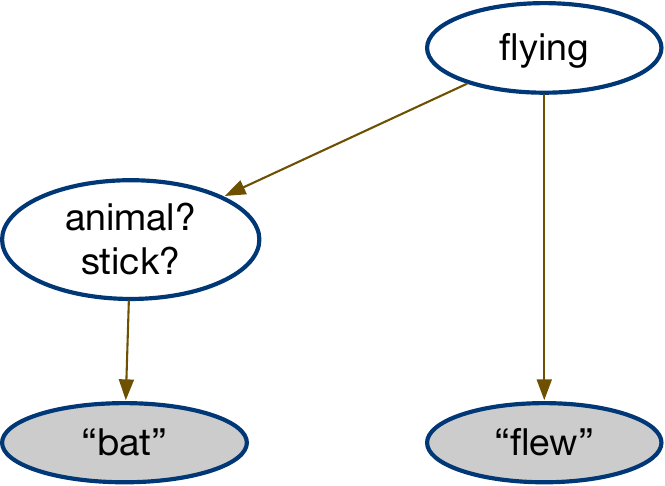}
  \caption{A graphical model that shows dependencies between random variables: the word \textit{bat} may have been uttered because the situation included some animal referent, or a baseball bat. Given the presence of the word \textit{fly}, the animal is more likely, though the baseball bat is not entirely impossible (as in \textit{The bat flew across the pitch}, presumably due to some player's frustration.) Shaded nodes are random variables whose values are known (called \emph{observed}).}
  \label{fig:bat_graphical}
\end{figure}

To start with, we will define the meaning of a word as a \emph{random variable}. A random variable is something that can take on different values, in a situation where we lack information about the true value. This lack of information is often described in terms of a random influence, for example the outcome of a coin flip. The range of values that the variable can take, and the probabilities associated with each single value, form a \textit{distribution}. For example, when we roll a fair six-sided die, the probability of getting any single number on the die is $\frac{1}{6}$. That is, the outcome of rolling the die is a random variable with distribution $\langle \frac{1}{6}, \frac{1}{6}, \frac{1}{6}, \frac{1}{6}, \frac{1}{6}, \frac{1}{6}\rangle$, indicating that there is a probability of $\frac{1}{6}$ to get a $1$, a $2$, a $3$, up to $6$. Note that the randomness refers to the outcome of the roll -- also known as \textit{sampling} -- not to the probability distribution. We can for instance imagine a loaded die with a distribution of $\langle \frac{1}{3}, \frac{1}{6}, \frac{1}{6}, \frac{1}{6}, \frac{1}{12}, \frac{1}{12}\rangle$, which constrains the outcome of the random process. Similarly, in a very simple model of word meaning, we can think of the word's semantics as being expressed by a distribution over word senses. The word \textit{bat} might have a $0.6$ probability of referring to a baseball bat, and a $0.4$ probability to refer to an animal, giving us the distribution $\langle0.6, 0.4\rangle$. Interpreting that word is akin to sampling a meaning according to a lexical die. The outcome of the roll is constrained by the particular shape of the die's probability distribution, which for a word, will reflect aspects we mentioned previously, such as selectional preference and discourse context.

The question becomes, then, how to express the interactions of different constraints. Following the random variable logic, we should also represent the probability of a certain type of event knowledge in a certain situation context, or the probability of a selectional preference given a specific verb in a specific lexical context. That is, we should have multiple random variables interacting with each other. To do this, we will talk of a a \emph{joint distribution}, which for e.g. three random variables $A$-$C$, we would write as $p(A \wedge B \wedge C)$. For instance, this could be the probability of a specific verb sense $A$ occurring with specific senses of that verb’s agent ($B$) in a situation type $C$. 

The dependencies between variables can be expressed as a \textit{graphical model}, as exemplified in Fig.~\ref{fig:bat_graphical}. This example shows a representation of the sentence \textit{A bat flew}, which has a natural interpretation as `an animal used its wings to move through the air', but could, given the right context, be understood as `a piece of sports equipment was flung through the air'. The grey nodes correspond to observed elements in the sentence (i.e. the words \textit{bat} and \textit{flew}), while the white nodes encode possible conceptual content, like a flying event. Arrows between nodes show dependencies. In this simple example, the uttered word \textit{bat} suggests the activation of either an animal concept or a sports equipment concept. The probabilities of those alternatives are dependent on the presence of some flying event in the situation, which in this case biases the interpretation towards the animal sense of \textit{bat}. The flying event itself can be inferred from having observed the word \textit{flew}.

\subsection{Interpretation as sampling process}

Given a graphical model, we can then explain how actual sentences are interpreted, via a so-called \textit{sampling} process. We can imagine navigating the graphical model from top to bottom, and at each node we encounter, throwing a loaded coin, or k-sided die, that reflect the probabilities of a random variable. For instance, using the example in  Fig.~\ref{fig:bat_graphical}, we might encounter the `animal / sports equipment' node and throw a loaded coin to sample either an animal concept or a sports equipment concept. The particular coin we throw is loaded in favor of animals if we have come from the \textit{flying} concept. Going further down the graph, we would throw another die to sample a word. Similarly, the die will be loaded in favour of words such as \textit{bird}, \textit{fly}, or \textit{bat} if we have come to the node from a path involving a flying event and an animal concept.

We can illustrate this process programmatically, by using the appropriate computational tool. WebPPL \citep{webppl} is a probabilistic programming language, with statements that have a probabilistic outcome. For example, it is possible to state
\begin{lstlisting}
var x = categorical({ps: [0.8, 0.2],
    vs: ["animal", "sports equipment"]})
\end{lstlisting}
to store in the variable \texttt{x} the outcome of a draw from a categorical distribution with probabilities (`ps') 0.8 and 0.2 for outcomes (`vs') `animal' and `sports equipment', respectively. Because we can program random draws, we can straightforwardly program the sampling process
`top down', sampling conceptual content and lexical items as we follow a particular path in the graph, and ending up with a full utterance. 

The second core ingredient in WebPPL is a command of the type
\begin{lstlisting}
  condition(sampled_predicate == utterance_predicate)
\end{lstlisting}
which rejects the sample that is currently in progress if the predicate we produce in the top-down process does not match the one that is observed in the utterance. For example, if we sample \textit{bird} from our animal concept, the sample is rejected because \textit{bird} does not occur in \textit{A bat flew}.

To sample multiple outcomes from the graphical model, we use a command of the type
\begin{lstlisting}
  var dist = Infer({method: 'rejection', samples: 2000,
    model:probfunction})
\end{lstlisting}
This call takes as one of its arguments a probabilistic function \verb+probfunction+, which it executes 2000 times using rejection sampling, recording each sample that finishes successfully, and storing the resulting empirical probabilities of samples in the variable \texttt{dist}. A sufficient large number of samples (here, 2000) must be used to get stable probability estimates. 

A probabilistic graphical model does not claim that the data was actually generated following the sampling process -- for the linguistic case, it does not claim that the speaker generates an utterance by sampling top-down to through the graph.
Rather, the graphical model
is only a convenient way of expressing the structure assumed by the probabilistic model when presented with the data, where the edges in the graph, the assumed dependencies, are the most important aspect.

\section{Situation description systems for word meaning in context}
\label{sec:situationdescriptions}

In this section, we formally define Situation Description Systems. Before we do that, we sketch some requirements for our definition.

\paragraph{A dual representation of meaning.} \citet{asher2011} assumes that a word is associated with both an intension and a type that is conceptual in nature. Similarly, we assume that a content word is associated with both an intension and with mental concepts. (In contrast to Asher, we assume concepts that belong to a particular listener, not listener-independent types.) Using a word brings to mind the associated concepts.


\begin{figure}[tb]
  \centering
  \includegraphics[scale=0.4]{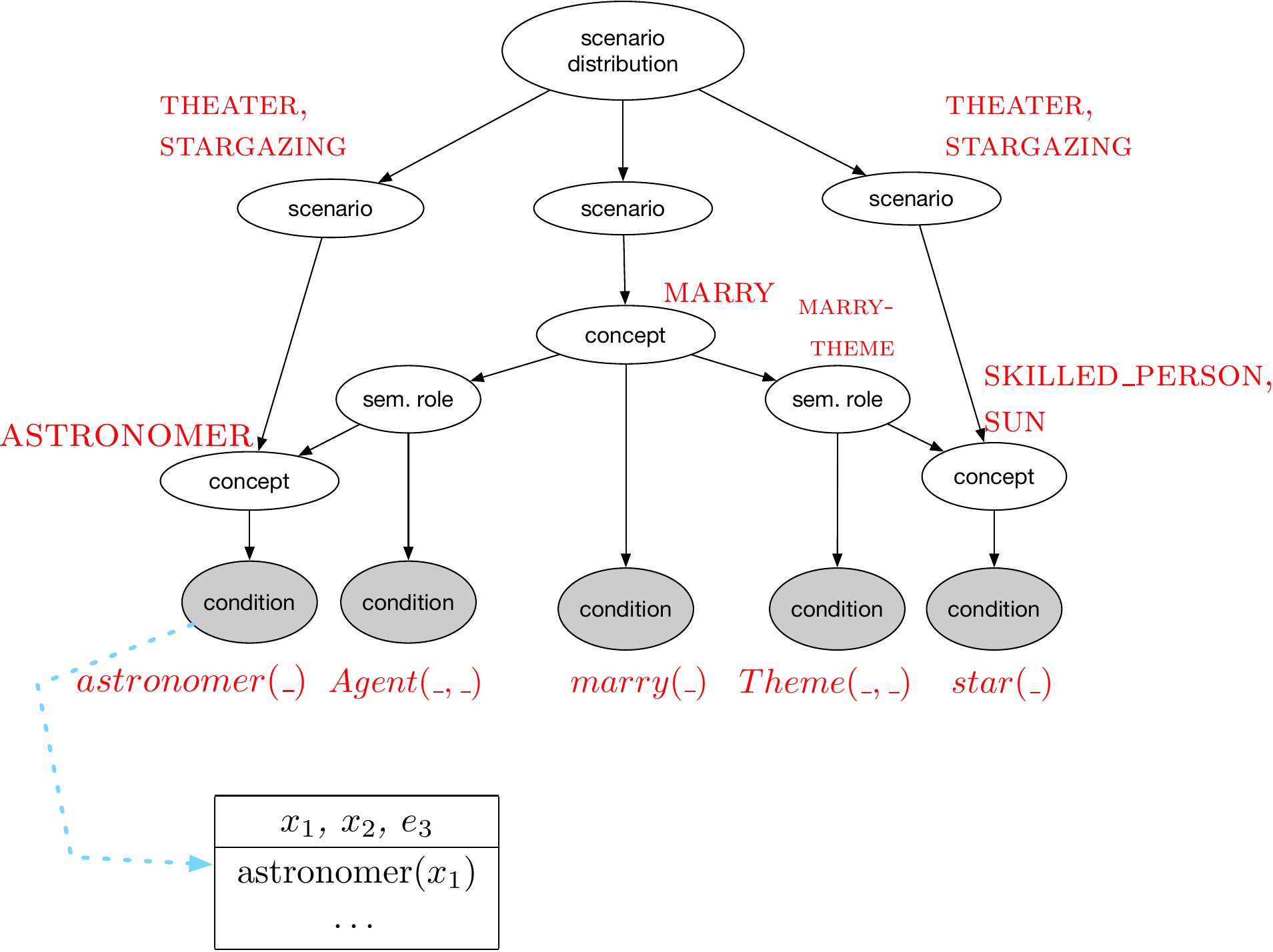}
  \caption{Directed graphical model and (part of) DRS for the sentence \textit{The astronomer married the star}. Nodes are random variables. In red: possible values for some random variables. Dashed blue line: Link between DRS condition and random variable.}
  \label{fig:astronomer_graphical}
\end{figure}

We use a two-part representation for the meaning of a sentence, with a Discourse Representation Structure (DRSs, \citet{KampReyle}) for the logical form, and a directed graphical model to represent the concepts underlying the words in the sentence. For the sentence above, the DRS is given in \ref{ex:astronomerdrs}.

\ex. \label{ex:astronomerdrs} \drs{$x_1$, $x_2$, $e_3$}{astronomer($x_1$)\\star($x_2$)\\marry($e_3$), Agent($e_3, x_1$), Theme($e_3, x_2$)}

Figure~\ref{fig:astronomer_graphical} shows the directed graphical model for this sentence.  The two parts of the sentence representation, the graphical model and the DRS, are connected through DRS conditions. For example, the condition $\text{astronomer}(x_1)$ in the DRS corresponds to a random variable whose value is $\textit{astronomer}(\_)$. In the figure, this correspondence is drawn as a blue arrow.

\paragraph{Interacting constraints, and uncertainty.} Many words have more than one concept with which they can be associated. For example \textit{star} can be linked to either the  \textsc{skilled\_person} or \textsc{sun} concept. (A \textit{star} can also be a star-shaped symbol on a keyboard, but we focus on just two concepts to simplify matters.) In a situation description, the sense of a word in a particular sentence context is indicated by its underlying concept. One primary goal of the formalization that we propose is to spell out the different constraints that influence word meaning in context. In the present paper, we focus on two types of constraints. First, we consider selectional constraints, for example that the concept \textsc{skilled\_person} is a better fit with \textsc{marry} than \textsc{sun} is. Second, we include scenario constraints. For example, the concepts \textsc{astronomer} and \textsc{sun} both fit a scenario we could call \textsc{stargazing}. 

We represent the possible concepts underlying \textit{star} as a random variable. In this case, the random variable has more than one possible value, as shown in red in Figure~\ref{fig:astronomer_graphical}. We also represent the scenario underlying each concept as a random variable (assuming that a sentence can draw on more than a single scenario). The constraints that influence the word meaning in context become edges in the graph. They need not express certainties but can be mere preferences. They can also interact and ``pull in different directions''. In fact, we will argue that this is what the selectional constraint and the scenario constraint do in this sentence, and this is why the sentence is perceived as a pun. 

\paragraph{Inference in graphical models.} In a graphical model, the values of some nodes may be known and fixed. These nodes are called \emph{observed}, and unobserved nodes are \emph{latent}. Inference in graphical models infers likely values for the latent nodes given the known values of observed nodes. In our case, DRS condition labels, like $\text{star}(\_)$, will be observed, shown as grayed nodes in Figure~\ref{fig:astronomer_graphical}. The values of all other nodes need to be inferred by the listener: From the words in the sentence, and from the way they are put together, the listener infers possible concepts and scenarios that may underlie the sentence.

\paragraph{Situation descriptions, and situation description systems.} We call the representation for a sentence -- a directed graphical model plus a DRS -- a \emph{situation description system}. A sample from the graphical model, i.e. an assignment of a value to each node, yields an individual \textit{situation description}. A given situation description represents one possible way of understanding the sentence, for instance with the concept \textsc{sun} underlying \textit{star}, and the scenario \textsc{stargazing} underlying that concept. Each situation description, because it is a sample from the graphical model, has an associated probability. When the listener feels that a word in the sentence has multiple senses that apply, we want the situation description system for the sentence to have multiple situation descriptions that have a probability greater than zero. 

\subsection{The probability space}
\label{sec:infinity}

Our situation description systems are probabilistic, so before we go any further we have to make sure that they do not run into any of the known problems with probability distributions over worlds or situations. \citet{Cooper:2015vj} argue that probability distributions over worlds  (which are used in \citealp{VanBenthem:2009te,vanEijckLappin,Zeevat:2013wc, Lassiter:adjectives} and \citealp{Erk:alligators}) are not cognitively plausible, and that neither are probability distributions over situations (as used by \citealp{Emerson} and \citealp{Bernardy:18}). We agree -- but, as we will argue, situation descriptions avoid the problems of  world distributions and situation distributions. 

A world is an unimaginably gigantic object. This is the reason why \citet{Cooper:2015vj} say it is unrealistic to assume that a cognizer could represent a whole world in their mind, let alone a distribution over worlds. A world is a maximal set of consistent propositions~\citep{Carnap}, and no matter the language in which the propositions are expressed, we cannot assume that a cognizer would be able to enumerate them. But the cognitive plausibility on which Cooper et al.\ focus is not the only problem. Another problem is that we do not know enough about a world as a mathematical object. \citet{Rescher:1999dq} argues that objects in the real world have an infinite number of properties, either actual or dispositional. This seems to imply that worlds can only be represented over an infinite-dimensional probability space. When defining a probability measure, it is highly desirable to use a finite-dimensional probability space -- but it is not clear whether that is possible with worlds. Maybe a world can be `compressed' into a finite-dimensional vector, but we simply do not know enough about worlds to say for certain. 

Situations, or partial worlds, may be smaller in size, but they still present similar problems, both in terms of cognitive plausibility and because they are underdefined. As reported above, \citet{Cooper:2015vj} make convincing arguments about plausibility aspects, so we concentrate on underdefinedness here. How large is, say, a situation where \emph{Zoe is playing a sonata}? Both \citet{Emerson} and \citet{Bernardy:18} assume, when defining a probability distribution over situations, that there is a given utterance (or set of utterances) and that the only entities and properties present in the situation are the ones that are explicitly mentioned in the utterance(s). But arguably, a sonata-playing situation should contain an entity filling some instrument role, even if it is not explicitly mentioned.
Going one step further, \citet{Clark:Bridging} discusses inferences that are ``an integral part of the message'', including bridging references such as ``I walked into the room. \textit{The windows} looked out to the bay.'' This raises the question of whether any situation containing a room would need to contain all the entities that are available for bridging references, including windows and even possibly a chandelier.  (Note that there is little agreement on which entities should count as available for bridging references: see \citealp{PoesioVieira:Bridging}.) The point is that there does not seem to be a fixed size that can be assumed for  the situation where \emph{Zoe is playing a sonata}.~\footnote{\citet{GoodmanLassiter} offer what looks like a way out of the problem of worlds. They assume that interpretation is always relative to a question under discussion, and that a world is generated only to the extent that it is relevant to the question under discussion. However, this still presumes enough knowledge about worlds to define a probability measure over them  in the first place -- but more importantly, questions under discussion do not have a fixed size any more than situations do.}

Our solution is to use a probability distribution over situation descriptions, which are objects in the mind of the listener rather than in some actual state-of-affairs. As human minds are finite in size, we can assume that each situation description only comprises a finite number of individuals, with a finite number of possible properties -- this addresses the problem that worlds are too huge to imagine. But we also assume that the size of situation descriptions is itself probabilistic rather than fixed, and may be learned by the listener through both situated experience and language exposure. Doing so, we remain agnostic about what might be pertinent for describing a particular situation.

\subsection{Uncertainty about situation descriptions}

We represent the meaning of an utterance as a \emph{distribution} over situation descriptions rather than a single best interpretation. There are several reasons for this choice. First, we want to be able to express uncertainty about the situation mentioned in the utterance. Going back to the example from above, \textit{Zoe was playing a sonata}, it is clear that some instrument must be involved, but not which instrument; and it is not clear what other participants or props are in the situation: a chair? a room? a teacher? other players? With a distribution over situation descriptions, we can express uncertainty about the number and nature of the participants and props in the situation.

Second, we want to be able to express uncertainty about the properties that can be inferred from a concept: If a particular discourse referent is an instance of \textsc{astronomer}, it is probably human, but not necessarily. Prevalent theories of concept representation in psychology (as reviewed for example in~\citealp{Murphy:02}) assume that there is no core set of necessary and sufficient properties that the instances of a concept share. We can model this by assuming that inferences apply probabilistically to concept instances, or in terms of a probabilistic graphical model, that a concept is endowed with probabilities of inferring different properties. Take, for example, the utterance \textit{Mary lied}. Following the analysis of \citet{ColemanKay:81}  of the verb \textit{to lie}, we could characterize the meaning of the utterance through multiple situation descriptions that differ in whether what Mary said was actually untrue, and whether she was intending to deceive.

Third, we have uncertainty about word sense. The astronomer sentence, repeated here as \ref{ex:astronomer_again}, is ambiguous between two very different senses of \textit{star}. Sentences can also be ambiguous with respect to word sense without being puns, for example sentence \ref{ex:checkboat}, reported by \citet{Hanks2000}. The sentence could be saying either that the man was inspecting the boat or stopping it. (He was stopping it, as the wider discourse context reveals.) We can describe \ref{ex:astronomer_again} by probabilistically sampling situation descriptions that differ in the concept underlying the word \textit{star}, and similarly with \ref{ex:checkboat} and \textit{check}. 

\ex. \a. \label{ex:astronomer_again} The astronomer married the \underline{star}.
\b. \label{ex:checkboat} Then the boat began to slow down. She saw that the man who owned it was hanging on to the side and \underline{checking} it each time it swung.

Our model uses collections of discrete concepts to model ambiguity, even in cases with multiple related meanings. \citet{hogeweg_vicente_2020} summarize the relevant work in psychology, where sense enumeration approaches to polysemy are broadly rejected, though there is no agreement on the best alternative. There does not seem to be a methodologically clean way to delimit concepts. But it greatly simplifies modeling to assume a collection of
discrete concepts, and in fact discrete concepts do not necessarily have to be far apart in meaning. This can be seen by taking clustering as a computational metaphor. There are computational clustering approaches that allow for overlapping or soft clusters, and often there are multiple clusterings of the data that yield very similar inferences. This computational metaphors matches our intuition of what concepts are and how the form; so we assume that we have distinct concepts that are not necessarily widely different in meaning. 

\subsection{Defining situation description systems}
\label{sec:situation_description_def}

We now formally define situation descriptions and situation description systems.

\paragraph{A DRT fragment.} A Discourse Representation Structure (DRS)  is a pair consisting of a set of \emph{discourse referents} $\{x_1, \ldots, x_n\}$  and a set of \emph{conditions} $\{C_1, \ldots, C_m\}$, written
\[\langle \{x_1, \ldots, x_n\}, \{C_1, \ldots, C_m\}\rangle\]
A DRS lists the discourse referents mentioned in a stretch of discourse along with the conditions that the discourse states about them. For example, if we assume a Neo-Davidsonian representation, with discourse referents for both entities and events, then we can represent the sentence \emph{a girl was sleeping}  as
\[\langle\: \{x, e\}, \{girl(x), sleep(e), Agent(e, x)\}\:\rangle\]
with the discourse referents $x, e$ and the conditions $girl(x), sleep(e), Agent(e, x)$ (if we ignore phenomena like tense and aspect). In the simplest case, a condition is an atomic formula, like $sleep(x)$, but a condition can also be a negated DRS $\neg D_1$ or it can be an implication $D_1\Rightarrow D_2$, where $D_1$ and $D_2$ are DRSs. 

In this paper we only consider a simple fragment of DRT, where the set of conditions consists of atomic formulas and negated atomic formulas, as this simplifies the link between the DRS and graphical model parts of the representation. We refer to this fragment of DRT as \emph{eDRT}, short for existential conjunctive DRT, as it cannot represent either universal quantifiers or disjunctions. We also only consider a simple fragment of the English language, of the form \textit{a Noun Verbed} or \textit{a Noun Verbed a Noun}, where the nouns denote objects, determiners are all indefinite, we only use singular forms and ignore phenomena like tense and aspect. For such sentences of English, the DRSs are actually eDRSs. We use a Neo-Davidsonian representation, with discourse referents for both entities and events, as exemplified above, and restrict ourselves to only unary and binary predicate symbols. 

Formally, eDRSs are defined as follows. Let $REF$ be a set of discourse referents, and $PS$ a finite set of predicate symbols with arities of either one or two. In the following definition, $x_i$ ranges over the set $REF$, and $F$ over the set of predicate symbols $PS$. The language of \emph{eDRSs} (existential and conjunctive DRSs) is defined by:

\begin{description}
\item[conditions] $C::= F(x) \mid \neg F(x) \mid F(x_1, x_2) \mid \neg F(x_1, x_2)$
\item[eDRSs] $D ::= (\{x_1, \ldots, x_n\}, \{C_1, \ldots, C_m\})$
\end{description}

We assume the standard model-theoretic interpretation for DRT. We further assume, for now, that the DRS for the utterance is fixed and has been built with one of the standard DRS construction methods, for example the top-down algorithm from \citet{KampReyle}. This assumption that the DRS is constructed ahead of time is a simplification, and  ignores interactions between structural ambiguity and word meaning ambiguity. It will be important to work out a semantic mechanism that combines DRS construction with probabilistic inference over concepts, but this goes beyond what we can do in the current paper.

\paragraph{Directed graphical models.}

A directed graphical model is a pair $M = (G, \Pi)$, where $G$ is a directed acyclic graph $G$. The nodes $X_1, \ldots,
X_n$ of the graph are random variables. The edges of the graph indicate dependencies among the random variables, where the value of each node $X_i$ depends on the values of its parent nodes in the graph. $\Pi = \{\pi_1, \ldots, \pi_n\}$ is the collection of these local conditional probability distributions, where $\pi_i$ is the conditional distribution for $X_i$. Writing $pa(X_i)$ for the parent nodes of the random variable $X_i$ in the graph, the probability of some assignment $A = a_1, \ldots, a_n$ of values to all the nodes in the graph is
\[ \prod_{i=1}^n P(X_i = a_i \mid pa(X_i), \pi_i)\]

Directed graphical models support different forms of inference. The one that is relevant to this paper uses observed random variables to infer values for all other (latent) random variables. In this case, the set of nodes $\{X_1, \ldots, X_n\}$ partitions into a set $\{Z_1, \ldots, Z_m\}$ of observed variables and a set $\{Y_{m+1}, \ldots, Y_n\}$ of latent variables. We have a partial assignment $A = a_1, \ldots, a_m$ of values to the observed variables, and we would like to know
\[P(Y_{m+1}, \ldots, Y_m \mid Z_1 = a_1, \ldots, Z_m = a_m)\]

Note that even though the edges are directed, they constrain both their adjacent nodes. We are not restricted to inferring values of child nodes given their parent nodes, we can also conversely infer likely values of parents given their children in a probabilistic analogue to modus tollens. In classical modus tollens, if we know all humans are mortal, but Zeus is not mortal, then we conclude that Zeus is not human. In its probabilistic analogue, if humans tend to live fewer than 150 years, but Zeus is over 1,000 years old, then Zeus is probably not human. To use a linguistic example, we turn to the sentence \textit{she drew a blade}, sentence \ref{ex:blade} from the introduction. Say the Theme of the concept \textsc{unsheathe} is most likely a weapon. In our framework we
would have a node for the concept underlying \textit{draw}, a node for the concept for \textit{blade}, and a node for the semantic role, with a directed edge from the \textit{draw}-node to the semantic role node, stating that the nature of the semantic role depends on the predicate, and a directed edge from the semantic role node to the \textit{blade}-node, stating the selectional constraint. Now say the concept underlying \textit{blade} is \textsc{blade\_of\_grass}, which is not a weapon. Then the semantic role is most likely not \textsc{unsheathe-Theme}, and hence the concept underlying \textit{draw} is most likely not \textsc{unsheathe}. 

\paragraph{Situation description systems.} We represent the meaning of sentences through situation description systems, of which Figure~\ref{fig:astronomer_graphical} shows an example. In the graphical model, we use eDRS condition labels, without their discourse referents, to indicate utterance words that have underlying concepts.  In the following definition, we write $label(c)$ for an eDRS condition without discourse referents: If  if $c = F(x)$ then $label(c) = F(\_)$, and if $c = \neg F(x)$ then $label(c) = \neg F(\_)$.

Intuitively, different eDRSs that only differ in variable names do not represent different meanings of a sentence.\footnote{Equivalence modulo variable renaming is simple in our current case, where we only consider one sentence at a time. In a dynamic DRS setting, variable renaming will have to be at the level of an agent's complete mental state.} Accordingly we first define proto situation description systems, which contain an eDRS, then the actual situation description systems, which contain equivalence classes over DRSs with respect to variable renaming. 

\vspace{2ex}
\begin{definition}[Proto situation description system]
  A proto situation description system is a tuple $(M, D, g)$ where $M$ is a directed graphical model, $D$ is an eDRS, and the function $g$ is a bijection between a subset of the random variables of $M$ and conditions of $D$. If $ g(X) = c$ for a random variable $X$ and condition $c$, then $label(c)$ is among the possible values of $X$. 

  Let the domain of $g$ be $dom(g) = \{Z_1, \ldots, Z_m\}$, and let $\{Y_{m+1}, \ldots, Y_n\}$  be the random variables of $M$ that are not in the domain of $g$. Then the  probability distribution of the proto situation description system is 

  \[P\big(Y_{m+1}, \ldots, Y_n \mid Z_1 = label(g(Z_1)), \ldots Z_m = label(g(Z_m))\big)\]
\end{definition}

This definition says that a proto situation description system comprises a directed graphical model and an eDRS. The two parts are linked by the function $g$, which ensures that the conditions in the eDRS are linked to random variables in the graph. The random variables $Z_k$ that have a $g$-mapping are the observed variables. If $Z_k$ is such a random variable, then its mapping $g(Z_k)$ is a condition in the DRS, and we stipulate that the observed value for $Z_k$ be the DRS condition label for its associated condition. 
For example, in Figure~\ref{fig:astronomer_graphical} the leftmost node in the bottom row has an observed value of $astronomer(\_)$. 

We group proto situation description systems into equivalence classes with respect to variable renaming in the eDRSs.
We say that two proto situation description systems $(M_1, D_1, g_1)$ and $(M_2, D_2, g_2)$ are equivalent if $M_1, M_2$ are identical, $D_1$ and $D_2$ are equivalent via a variable mapping $f$, and  for any node $X$ of $D_1$ it holds that $X  \in dom(g_1)$ iff $X \in dom(g_2)$, and  if $X \in dom(g_1)$ then $f(g_1(X)) = g_2(X)$.  

\vspace{2ex}
\begin{definition}[Situation description system]
A \emph{situation description system} is an equivalence class of proto situation description systems with respect to eDRS variable renaming. 
\end{definition}

Abusing notation, we write $(M, D, g)$ for the equivalence class containing the proto situation description system $(M, D, g)$ whenever there is no danger of confusion.

\paragraph{Situation descriptions.} A situation description is one possible interpretation of a sentence. Formally, it is an assignment of values to all nodes of the graphical model in the situation description system that respects the observed node values. 

\vspace{2ex}
\begin{definition}[Situation description]
A situation description is a tuple $(M, D, g, A)$ where $(M, D, g)$ is a situation description system, and $A$ is an assignment of values to the nodes of $M$ such that for any node $X \in dom(g)$, $A(X) = label(g(X))$. 
\end{definition}

As discussed in \S~\ref{sec:infinity}, situation description systems are mental representations that are finite in nature, and we want to ensure that they cannot grow arbitrarily large. The directed graphical models that we construct in the next section are bounded in size by the number of discourse referents in their  associated eDRSs. So if we assume that there is some upper limit on the number of discourse referents, our situation description systems are bounded in size.

\section{Situation description systems for word meaning in
  context: Selectional constraints only}
\label{sec:sds_selpref}

In this and the following sections we show how to use situation description systems (SDSs)
to 
specify constraints on word meaning in context. We start with SDSs
that only use selectional constraints. For now, all nodes in the
graphical model have values that are either concepts or semantic
roles, so we should briefly comment on what we mean by concepts and
semantic roles. We assume that ``word meanings are
made up of  pieces of conceptual structure''~\citep[p.391]{Murphy:02}. So at a
coarse-grained level,  a word occurrence evokes a \emph{concept}, which
disambiguates the word. 
\emph{Semantic roles} characterize the participants of
events by the way they are involved. There are many proposals which
spell out the set of semantic roles one should assume, as well as
their level of granularity. We do 
not need to commit to any 
particular semantic role inventory, it is enough to assume that there
is some overall finite set of semantic roles, and that some
concepts are associated with semantic roles, with role labels that
could be specific to the concept or general across concepts. Here, we
use roles that are specific to their concept, like
\textsc{sleep-theme}, so we do
not have to engage with the question of the granularity of semantic
roles. 
We assume that semantic roles
characterize the goodness of role fillers through selectional
constraints. Following models from psycholinguistics, including
\citet{McRaeThemrolesAsConcepts} and \citet{Pado:09} we assume these
constraints are often selectional \emph{preferences} rather than
hard constraints.

\begin{figure}[tb]
  \centering
  \includegraphics[scale=0.4]{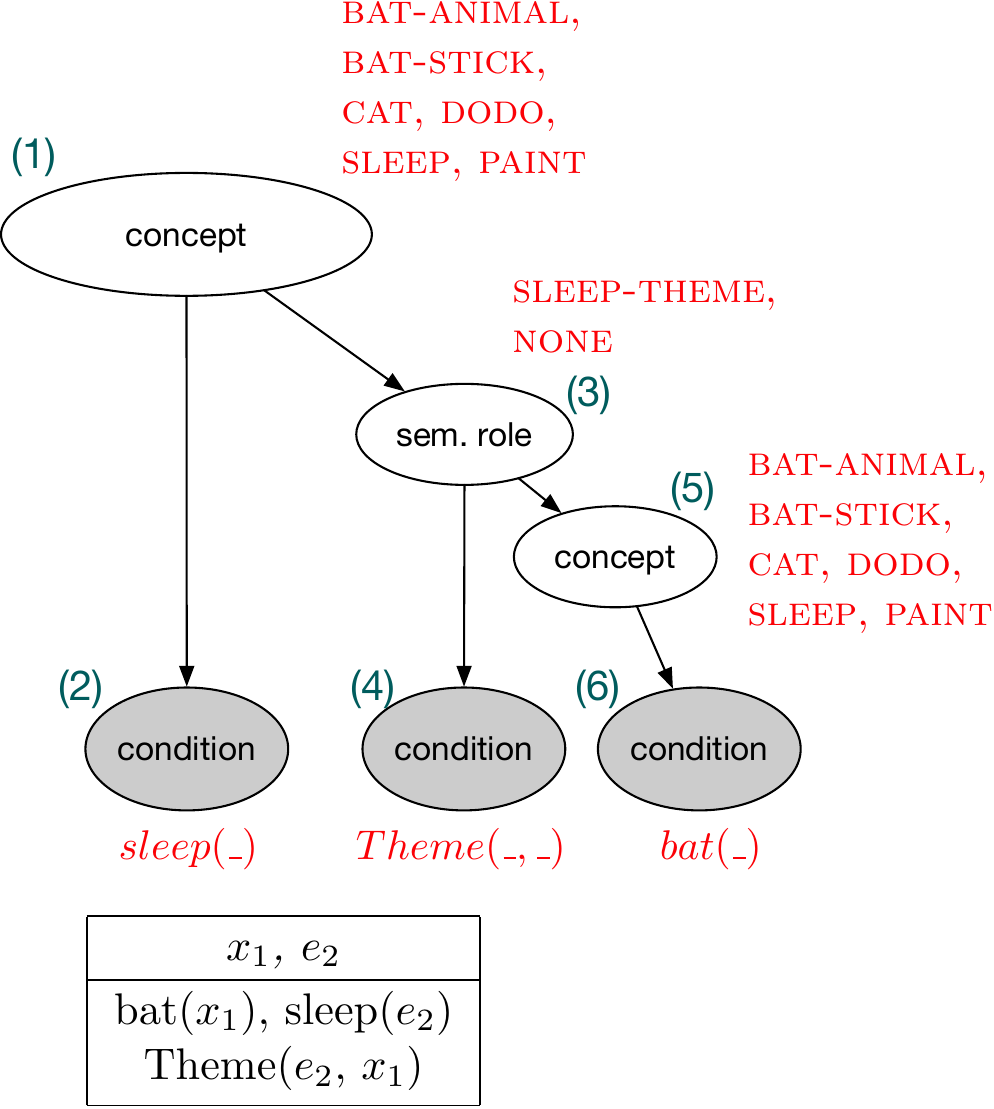}
  \caption{Situation Description System for the sentence \textit{A bat was
      sleeping}, with selectional constraints only. For each random variable, the list of all possible
    values is shown in
    red. Node numbers have been added for easier discussion in the
    text.}
  \label{fig:selpref_bayesnet}
\end{figure}

\subsection{A simple example without ambiguity}

Figure~\ref{fig:selpref_bayesnet} shows an SDS for 
sentence \ref{ex:selpref_sleep}, using selectional
constraints. We need to ascertain that this SDS licenses the right
situation descriptions, with probabilities that intuitively make
sense. As a situation description is a sample from the graphical
model, we do this by going through the sampling process one node at
a time from the top down.

\ex. \label{ex:selpref_sleep} A bat was sleeping.

\paragraph{Nodes (1) and (2): the concept underlying \textit{sleeps},
  and an observed condition label.} 
We start at the top node (1) in
Figure~\ref{fig:selpref_bayesnet}. This is a random variable that
ranges over concepts. In the example in Figure~\ref{fig:selpref_bayesnet}, we assume a
rather limited concept inventory in order to keep things simple,
comprising only the concepts \textsc{bat-animal}, \textsc{cat},
\textsc{dodo}, \textsc{bat-stick},
\textsc{sleep}, and \textsc{paint}.

The probability distribution associated with  node
(1) is a \emph{multinomial distribution}, a distribution over
categorical values, in this case, concepts. The textbook example for multinomial
distributions is rolling a die. With a fair six-sided die and a single
throw, the
probability of each side is $\frac{1}{6}$. For a trick die, the
probabilities could also vary, for example $\frac{1}{3}$ for rolling a
one but only $\frac{1}{12}$ each for rolling either a five or a six. We have an inventory of six
concepts, and if they are all equally likely, each one of them, too,
has a probability of $\frac{1}{6}$. 

The probability distribution for node (2) is also a multinomial distribution, this time with values that are labels like
sleep$(\_)$, bat$(\_)$, Theme$(\_, \_)$. This is a conditional probability
distribution, dependent on the value of the parent node (1). That is,
for each possible value $c$ of node (1), node (2) has a multinomial
distribution $P(value | c)$.

Node (2) is an observed node, we
already know what its value is, namely sleep$(\_)$. Given our
extremely limited concept inventory, it makes sense to assume that
there is only one single value $c$ of node (1) for which
$P(\text{sleep}(\_) \mid c)$ is non-zero, namely $c = $
\textsc{sleep}. Then by probabilistic modus tollens, in any sampled SD
that has non-zero probability the concept in
node (1) has to be \textsc{sleep} .

\paragraph{Nodes (3) and (4): a semantic role.} For a semantic
role, we use a random variable with two possible values: the role
exists and is filled in the situation, or not. The associated
probability distribution is a \emph{Bernoulli distribution}. The
standard textbook example of this distribution is the coin flip, with
some probability $p$ of the coin coming up heads, and $1-p$ of it
coming up tails. The probability 
distribution associated witih node (3) is again a 
conditional distribution, dependent on the value of the parent node
(1): Given that the parent node's value is the concept \textsc{sleep},
how likely is the concept \textsc{sleep} to have a \textsc{theme}
present in the described situation?

Node (4) is observed, with a value of Theme$(\_,
\_)$. Since it conditionally depends on node (3),
the probabilistic equivalent of modus tollens again gives us that the
value of node (3) must be \textsc{sleep-theme} rather than
\textsc{none}. 

In our case, the
probability $P(\text{\textsc{sleep-theme}} \mid
\text{\textsc{sleep}})$ is the probability that the role
\textsc{sleep-theme} would be realized for any occurrence of
\textsc{sleep}. It makes sense to assume that the sleeper always needs to be realized
in a sleeping event, that is, $P(\text{\textsc{sleep-theme}} \mid
\text{\textsc{sleep}}) = 1$, and the probability of the role not being
realized is $P(\text{\textsc{none}} \mid
\text{\textsc{sleep}}) = 0$. 

The observed value $Theme(\_, \_)$ of node (4) is conditionally
dependent on node (3): If the value of node (3) had been
\textsc{none}, the value of node (4) would also be $None$,
unrealized.\footnote{There are other semantic role nodes not shown in the picture, one
for each possible semantic role in the system. In this case there are 
\textsc{paint-agent} and \textsc{paint-theme}, both with zero
probability of occurring as roles of \textsc{sleep}.}

\paragraph{Nodes (5) and (6): the concept underlying \textit{bat}, and
  associated condition label.} Node (5), like node (1), is a random
variable whose values are concepts. It is associated with a
multinomial distribution, but this time it is a conditional
distribution dependent on node (5). This is the distribution that
characterizes the selectional preference: Given that the semantic role
is \textsc{sleep-theme}, how good a role filler is an instance of a  \textsc{cat} or
\textsc{bat-animal} as opposed to an instance of a \textsc{bat-stick}?
To express that only animate entities sleep, we can set
\[
P(c \mid \text{\textsc{sleep-theme}}) = \left\{\begin{array}{ll}
                                    \frac{1}{3} & \text{for }c \in \{\text{\textsc{bat-animal}}, \text{\textsc{cat}},  \text{\textsc{dodo}}\}\\
                                    0 & \text{else}
\end{array}\right.
\]                                             

In this paper, we specify 
probabilities by hand in order to demonstrate how an SDS works. To
scale up, probabilities will need to learned automatically from
data. This can be done by using FrameNet
frames~\citep{Fillmore:Framenet} as concepts, doing 
automatic frame-semantic 
parsing~\citep{xia-etal-2021-lome} on corpus data and counting co-occurrences of the
resulting frames. 

The only concepts  with non-zero probability under the conditional
distribution $P(c \mid \text{\textsc{sleep-theme}})$ are
\textsc{bat-animal}, \textsc{cat},  and \textsc{dodo}. 
In addition, the observed value of node (6) is $bat(\_)$, and $P(bat(\_) \mid c)$ is zero for all concepts $c$
except for \textsc{bat-animal} and \textsc{bat-stick} -- but 
\textsc{bat-stick} has been eliminated by $P(c \mid
\text{\textsc{sleep-theme}})$, leaving only  \textsc{bat-animal}.

\subsection{Examples with ambiguity}

\begin{figure}[tb]
  \centering
  \includegraphics[scale=0.28]{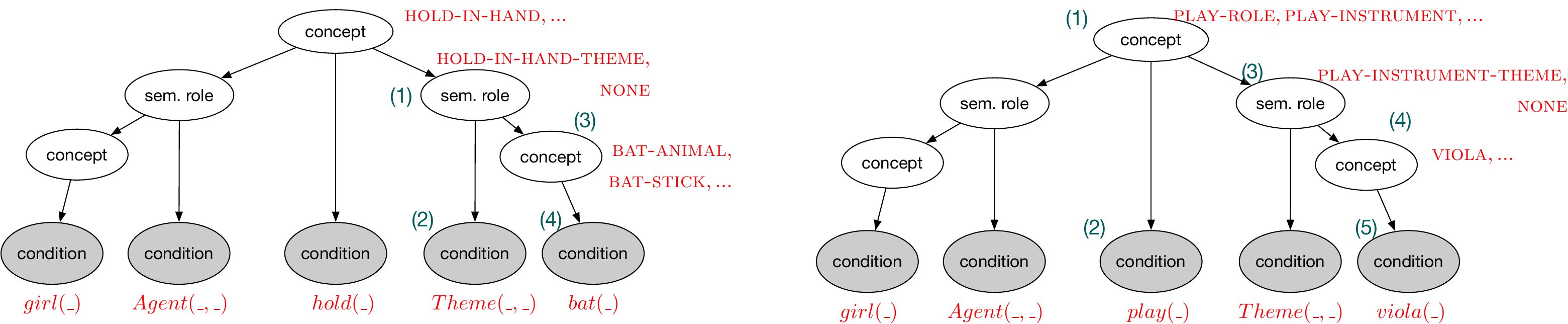}
  \caption{Graphical model part of the SDS for \textit{A girl was holding
      a bat} (left) and \textit{a girl was playing a viola}
    (right). Some possible values of random variables shown in red. Node numbers have been added for easier discussion in the
    text.}
  \label{fig:selpref_amb}
\end{figure}

We next consider two cases that arguably allow for more than a single
interpretation, in \ref{ex:selpref_amb1} and \ref{ex:selpref_amb2},
with the graphical models of their SDSs sketched in Figure~\ref{fig:selpref_amb}. 

\ex. \label{ex:selpref_amb1} A girl was holding a bat.

\ex. \label{ex:selpref_amb2} A girl was playing a viola. 

\paragraph{The sense of \textit{bat} in sentence
  \ref{ex:selpref_amb1}.} In the left sentence of
Figure~\ref{fig:selpref_amb}, the node
of interest is node (3), the concept underlying \textit{bat}. Node (1) is a semantic role node whose
value must be \textsc{hold-in-hand-theme} rather than \textsc{none} because of
observed node (2). Node (3) is conditionally dependent on node (1),
and the conditional distribution again constitutes the selectional
preference of \textsc{hold-in-hand-theme}. For simplicity, assume that any
concept describing concrete objects is equally likely as a Theme of
\textsc{hold-in-hand}, so \textsc{bat-animal} and \textsc{bat-stick}
have the same probability of appearing as \textsc{hold-in-hand-Theme}. However, node (4) has an observed value of
bat$(\_)$. If we assume that \textsc{bat-animal} and
\textsc{bat-stick} are the only concepts that can sample that label,
then by our probabilistic analogue of modus tollens, they are the only
possible values for node (3). We obtain two different situation
descriptions with equal probability, one in which a girl is holding a
stick, and one in which a girl is holding an animal. 

\paragraph{The sense of \textit{play} in sentence
  \ref{ex:selpref_amb2}.} In the right sentence of
Figure~\ref{fig:selpref_amb}, the node
of interest is node (1), the concept underlying \textit{play}. Assume
that there are exactly two concepts $c$ for which
$P(\text{play}(\_)\mid c)$ is non-zero, and which therefore match the
observed value of node (2): \textsc{play-role} and
\textsc{play-instrument}. 
Say we sample the value of node (1) to be
\textsc{play-instrument}. Then the only roles with non-zero probability
are \textsc{play-instrument-agent} and
\textsc{play-instrument-theme}. Since the observed label in node (4) is
Theme$(\_, \_)$, node (3) must have sampled that the
role \textsc{play-instrument-theme} is present rather than absent. We
can assume the Theme of \textsc{play-instrument} to have a very
specific selectional preference: The role has to be filled by an
instrument. For node (5), the filler of the role, it makes sense to
assume that \textsc{viola} is the only concept from which the observed
value viola$(\_)$ of node (6) can be sampled. The concept
\textsc{viola} is a good filler for the Theme.

Now assume that instead we sample \textsc{play-role} for node (1). In
this case, the only roles with non-zero probability are
\textsc{play-role-agent} and \textsc{play-role-theme}. Again,
\textsc{play-role-theme} must be sampled to be present rather than
absent because of the observed Theme$(\_, \_)$. But for the Theme of
\textsc{play-role}, \textit{viola} is not as good a filler as, say,
\textit{prince}. We first look at the case in which this is a hard
constraint, and the Theme has to be animate. In that case,
$P(\text{\textsc{viola}}\mid \text{\textsc{play-role-theme}}) = 0$, so
by probabilistic modus tollens, the value of node (4) cannot be
\textsc{play-role-theme}, so the value of node (1) cannot be
\textsc{play-role}. So if we set up the selectional constraint of
\textsc{play-role} to be a hard constraint, the word \textit{viola}
disambiguates \textit{play} to mean play an instrument rather than
play a role.

We can also make the selectional constraint \textsc{play-role-theme}
soft. In that case, we obtain two different SDs with non-zero
probability, one in which a girl makes music on a viola, and one in
which a girl plays the role of a viola, maybe in a theater performance.  To see how
the probabilities work in practice, we use computational
simulations of the sampling 
process. As discussed in Section~\ref{sec:generative}, probabilistic
programming languages such as WebPPL~\citep{webppl} allow us to
directly implement probabilistic graphical models, such as SDSs, and
draw samples. We do this for our sentence \ref{ex:selpref_amb2}. We
use a reasonably high number of samples, 2,000, such that each SD will be
sampled multiple times, giving us reasonable
empirical probabilities (simulation-based estimates of the theoretical
probabilities) of the different possible concepts underlying
\textit{play}. 

We assume the following concept inventory:
\begin{quote}
  \textsc{play-instrument}, \textsc{play-role}, \textsc{viola},
  \textsc{cello}, \textsc{girl}, \textsc{prince}, \textsc{dodo}, \textsc{sheep}
\end{quote}

We require the Theme of \textsc{play-instrument} to be an instrument,
with probabilities of 0,5 each for \textsc{cello} and \textsc{viola}
and zero otherwise. We make the selectional constraint for the Theme
of \textsc{play-role} soft, with
\[P(c \mid \text{\textsc{play-role-theme}}) = \left\{\begin{array}{ll}
                                    0.24 & \text{for }c \in
                                           \{\text{\textsc{girl}},
                                           \text{\textsc{prince}},
                                           \text{\textsc{dodo}}, \text{\textsc{sheep}}\}\\
                                                       0.02 &
                                                              \text{for
                                                              } c \in
                                                              \{\text{\textsc{viola}},
                                                              \text{\textsc{cello}}\}\\
                                                       0 & \text{else}\\
\end{array}\right.
\]

The chosen probabilities are to some extent arbitrary, they simply
reflect a high probability for animate fillers, and a low but nonzero
probability for other concepts of concrete objects. With this particular choice of probabilities, our WebPPL model gives us empirical probabilities of 0.98 for
\textit{play} meaning \textsc{play-instrument}, and 0.02 for
\textsc{play-role}. 

\section{Situation description systems for modeling word meaning in
  context: Selectional constraints and scenarios}
\label{sec:sds_scenario}

We now add scenario constraints to our inventory. By a \emph{scenario}
we mean knowledge about the world that can involve groups of events and
entities that often appear together. Such world knowledge is discussed in psychology
under the name of \emph{generalized event
knowledge}~\citep{McRae:LLC}, and in artificial intelligence as
\textit{scripts} or \textit{narrative schemas}~\citep{schank_abelson77,Chambers:08}. \citet[p.111]{Fillmore1982frame}
describes word meanings in terms of frames that words evoke,
characterizing frames as ``a general cover term for [\ldots] ‘schema’,
‘script’, ‘scenario’ \ldots 'cognitive model.'' Like \citet{McRae:LLC} and
\citet[p.130]{Fillmore1982frame}, we assume that listeners expect an
utterance to more often than not describe a coherent scenario -- though
it is of course possible for an utterance to tap into multiple
scenarios. 


We posit that it is the content words in a sentence that
link to scenarios,\footnote{FrameNet also has frames evoked by, for
  example, prepositions,   multi-word expressions, and larger
  constructions, but we restrict ourselves to content words here for
  simplicity.} or more precisely, that the \textit{underlying concept} of a
word links to a scenario. We further assume that each concept in the sentence
is linked to an underlying scenario. Scenarios can be repeated, if several concepts are generated by the same script, or they can take different values.


In the formalization below, we describe selectional constraints as
specific to mental concepts, but influenced by the scenario(s) that
are active in the sentence; this is consistent with the evidence
presented in \citet{SanfordGarrod}.

\subsection{Disambiguation by scenario}

For probabilistic modeling, there is a convenient framework that
encodes exactly the kind of structure we need, called topic modeling
or Latent Dirichlet Allocation~\citep{Blei:03}. This probabilistic
framework has been used to characterize topics in a document, or
senses of a word~\citep{dinu2010}.\footnote{\citet{Ferraro:2016} integrate scripts  and frames with a Latent   Dirichlet Allocation variant. Their framework focuses on   end-to-end learnability of the probability distribution and therefore compromises on linguistic expressivity. We go the opposite route. We focus on the ability of the framework to express linguistic phenomena as accurately as possible, even though our framework may not allow for end-to-end learning.} Topic modeling gives us two key ideas for situation descriptions. The first idea formalizes the notion of a scenario as a group of
events and entities that belong together: Each scenario is associated with a multinomial
distribution over concepts. For example, a
\textsc{sports} scenario might give high probability to concepts such
as \textsc{player}, \textsc{ball} or \textsc{bat-stick}.\footnote{This is a simplification of scenarios. A more   sophisticated representation, like the one used by \citet{chersoni_santus_pannitto_lenci_blache_huang_2019},  would take into account that concepts
  tend to appear in scenarios in specific roles, for example a vampire in
  a gothic scenario would be a typical attacker. But for now we keep
  our representation as simple as possible.} The second
idea formalizes the notion that an utterance draws on one or more
scenarios, i.e. each utterance is associated with an underlying multinomial distribution
over scenarios. For each content word, a scenario is sampled from this
multinomial, then a concept is sampled from the scenario. If the
scenario distribution of the utterance is sparse, that is, it only has
one or a few scenarios that get non-zero probabilities, then this
multinomial implements our intuition that utterances are typically
coherent in their scenario(s). We show how this works in detail using
sentence \ref{ex:ambiguous}, repeated here as \ref{ex:playerbat}. 

\ex. \label{ex:playerbat} A player was holding a bat. 

The sentence is ambiguous between the \textit{animal} and the \textit{stick} sense
of \textit{bat}, where arguably the \textit{stick} sense is much
more prominent. This difference in prominence cannot come from the selectional
constraint, as both animals and sticks can be held equally well. But
we can explain the preference for the \textit{stick} sense as a
preference of the listener to repeatedly draw on the same scenario,
and both \textit{player} and the \textit{stick} sense of \textit{bat}
fit well into a \textsc{sports} scenario.

\begin{figure}[tb]
  \centering
  \includegraphics[scale=0.4]{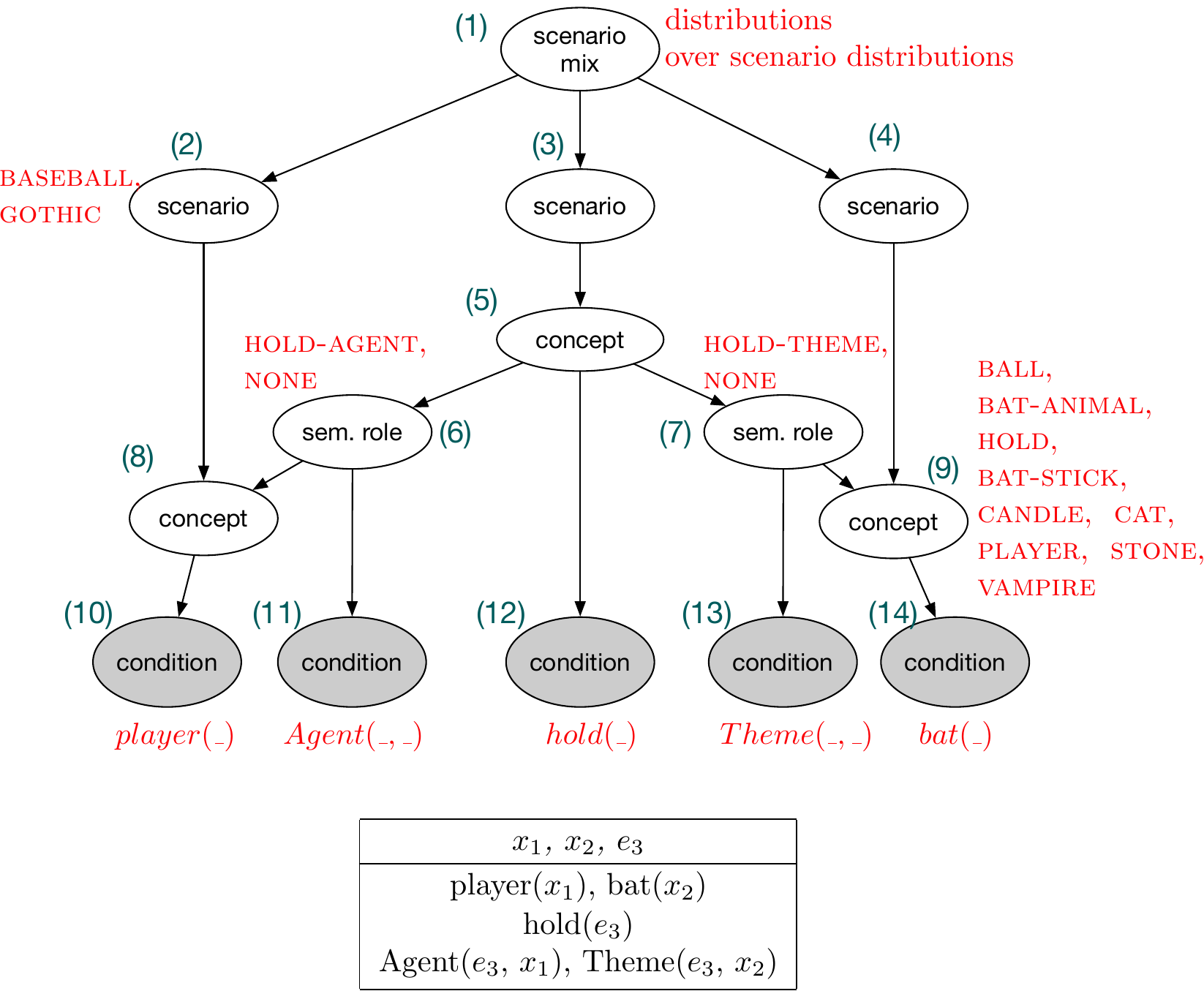}
  \caption{Situation Description System for the sentence \textit{A
      player was holding a bat}, with both selectional constraints and
    scenario constraints, For each random variable, the list of all possible
    values is shown in
    red. Node numbers have been added for easier discussion in the
    text.}
  \label{fig:scenario_sds}
\end{figure}

Figure~\ref{fig:scenario_sds} shows the SDS for
sentence~\ref{ex:playerbat}, with a graphical model that is larger
than before because it now also contains random variables for
scenarios. Here is how a situation description would get sampled from
this SDS. 

\paragraph{Node (1): The scenario distribution for the sentence.}
The sentence as a whole is associated with a distribution over
scenarios. There is not a single scenario distribution that holds for
all sentences, instead the listener infers, upon hearing the sentence,
which scenarios the sentence is about. For this reason, the
distribution over scenarios gets its own random variable, node
(1). That means that any value for node (1) is a distribution over
scenarios, and the distribution from which that value is sampled is
hence a distribution over multinomial distributions.

Concretely, say our listener is only aware of two 
scenarios, \textsc{baseball} and \textsc{gothic}. Then the possible
values for node (1) include 
$\langle$\textsc{baseball}:0.9, \textsc{gothic}:0.1$\rangle$, and
$\langle$\textsc{baseball}:0.5, \textsc{gothic}:0.5$\rangle$. Such
values can be sampled from a \emph{Dirichlet
  distribution}, which is, in fact, a distribution over
multinomial distributions.  

A multinomial distribution has as its parameters the probabilities it
gives to each category, for example $\langle$\textsc{baseball}:0.9,
\textsc{gothic}:0.1$\rangle$. A Dirichlet distribution has as its
parameter the \emph{concentration parameter} $\alpha$, which
influences what kinds of multinomials we are likely to sample from
it. Conveniently, when $\alpha$ is smaller than one, then the
Dirichlet distribution is
more likely to produce sparse multinomial distributions: It is more
likely to sample $\langle 0.9, 0.1\rangle$ or
$\langle 0.02, 0.98\rangle$ than $\langle 0.5, 0.5\rangle$. This is
exactly what we want: By choosing an $\alpha <1$, we can impose the
constraint that utterances are typically
coherent and avoid `mixing' scenarios.

\paragraph{Node (3): the scenario underlying \textit{holds}.} 
Node (3) is a random variable whose values are scenarios, in our case
\textsc{baseball} and \textsc{gothic}. It is conditionally dependent
on node (1), the scenario distribution for the sentence.
If, for example, the value of node (1) is $\langle$\textsc{baseball}:0.9,
\textsc{gothic}:0.1$\rangle$, then node (3) samples its value from
that multinomial, and its value is most likely to be \textsc{baseball}.

Each scenario \textsc{baseball} is associated with a multinomial
distribution over concepts, to express the events and entities that
typically appear in the scenario. For the sake of illustration, we
manually set those distributions as follows. We assume a small
inventory of concepts: 
\begin{quote}
\textsc{ball}, \textsc{bat-animal}, \textsc{hold},
\textsc{bat-stick}, \textsc{candle}, \textsc{cat}, \textsc{player}, \textsc{stone},  \textsc{vampire}
\end{quote}
We set the multinomial distribution for the \textsc{baseball}
scenario to give equal probability 
to the concepts \textsc{ball}, \textsc{bat-stick}, \textsc{hold},
\textsc{player}, and \textsc{stone}, and zero probability
to all other concepts. We set the distribution  for  \textsc{gothic} to give equal
probability to the concepts \textsc{bat-animal}, \textsc{candle},
\textsc{cat},  \textsc{hold}, and \textsc{vampire}, and zero
probability otherwise.

Again, we set all probabilities by hand in this
paper; to scale up, scenario probabilities could be learned by applying topic modeling
on top of a corpus that has been automatically labeled with frames.

\paragraph{Node (5): the concept underlying \textit{holds}.}
Node (5) is a random variable that stands for the concept underlying
\textit{holds}, and now that we have scenarios, node (5) is
conditionally dependent on node (3). If the value of node (3) is
\textsc{baseball}, then the value for node (5) gets sampled from the
multinomial distribution for the \textsc{baseball} scenario. As node
(12) is observed with a value of hold$(\_)$, and with the assumption
that \textsc{hold} is the only concept from which this label can be
sampled, the concept in (5) has to be \textsc{hold}.

\paragraph{Node (7): a semantic role.} The concept \textsc{hold} has
two roles, \textsc{hold-agent} and \textsc{hold-theme}. The value of
node (7) has to be \textsc{hold-theme} rather than \textsc{none}, as
in the examples before. The role \textsc{hold-theme} is again associated with a selectional
constraint, which is expressed as a multinomial distribution over
concepts. We manually set the selectional preference to allow for all
concrete objects, so 
\[
  P(c \mid \text{\textsc{hold-theme}} = \left \{ \begin{array}{ll}
                                             0 & \text{for } c = \textsc{hold}\\
                                             0.125 & \text{else}\\
\end{array}\right.
\]

\paragraph{Node (9): the concept underlying \textit{bat}.} The random
variable in node (9) characterizes the concept underlying
\textit{bat}. It is conditionally dependent not only on node (7), the
semantic role, but also on node (4), its underlying scenario. Both the
semantic role and the scenario express their preferences through a
multinomial distribution over concepts. We want to say that the role
filler is chosen to match \emph{both} constraints, so we combine the two
distributions using a \emph{Product of
  Experts}.

In general, a Product of Experts implements the following process. Say we have
two multinomial distributions over $k$ categories, with
event probabilities $a_1, \ldots, a_k$ and $b_1, \ldots, b_k$
respectively. Then the product-of-experts probability of class $i$ is
\[\frac{a_i b_i}{\sum_j a_jb_j}\]
The numerator is the probability that both distributions ``vote
for'' $i$, and the denominator is the sum of the probabilities of
all outcomes where both
distributions agree on their ``vote''.
\footnote{It could happen that the scenario and
the selectional preference do not
have any concept that they could agree on -- that is, that there is no
concept to which they both give a non-zero probability. In that
case the Product of Experts distribution would not be not
well-defined. The
  overall probability distribution would become ill-defined if, for
  example, every SD one could sample would involve a required role that
  cannot be  filled at all. It is possible to adapt all formulas to explicitly
account for such pathological 
cases. We are not showing the adaptation here in order to keep the formulation simple, but we 
address these cases in the implementation.}

Because of the observed value of $bat(\_)$ in node (14), the value of
node (9) has to be either \textsc{bat-animal} or \textsc{bat-stick} by
probabilistic modus tollens. The selectional constraint of
\textsc{hold-theme} does not have a preference for either sense of
\textit{bat} over the other. So the choice of concept depends on the
scenario: If the scenario in node (4) is \textsc{baseball}, then we
are more likely to sample \textsc{bat-stick} in node (9), and if the
scenario in node (4) is \textsc{gothic}, then node (9) is more likely
to be \textsc{bat-animal}. To see which value of node (4) is more
likely, we have to look to nodes (2) and (8).

\paragraph{An inference walk through the whole tree, starting at node
  (8), the concept underlying
  \textit{player}.}
Node (10) has an observed value of $player(\_)$, so the concept sampled for node (8) has to be
\textsc{player}. This value is conditionally dependent on nodes (2) and (6), the
scenario and semantic role. So by probabilistic modus tollens the 
scenario sampled for node (2) has to be \textsc{baseball}, as
\textsc{gothic} gives zero probability to \textsc{player}. 
But if the
scenario in node (2) is \textsc{baseball}, then again by
probabilistic modus tollens, node (1) is more likely to have a value
like $\langle$\textsc{baseball}:0.9,
\textsc{gothic}:0.1$\rangle$ than a value like $\langle$\textsc{baseball}:0.1,
\textsc{gothic}:0.9$\rangle$. (And if the Dirichlet
distribution in node (1) prefers sparse distributions, a value like $\langle$\textsc{baseball}:0.5,
\textsc{gothic}:0.5$\rangle$ is unlikely in general). And if that is so, then node (4) is
more likely to have a value of \textsc{baseball} than
\textsc{gothic}. Finally, if node (4) is likely to be \textsc{baseball},
then node (8) is likely to be \textsc{bat-stick}.

\begin{table}[tb]
  \centering
  \begin{tabular}{l|ll}
    setting& p(stick) & p(animal)\\\hline
    selectional constraints only & 0.50 & 0.50\\
    with scenario constraintss, $\alpha = 0.5$ & 0.82 & 0.18\\
    with scenario constraints, $\alpha = 0.1$ & 0.96 & 0.04\\
  \end{tabular}
  \caption{Empirical probabilities for the "stick" and "animal" senses
    of \textit{bat} in "A player was holding a bat", with only
    selectional constraints, or with both selectional constraints and
    scenario constraints (WebPPL simulation)}
  \label{tab:player_bat}
\end{table}

\paragraph{Experimentally testing the sentence representation.} We
again use a computational simulation to see what probabilities we get
for the different senses of \textit{bat} in the sentence, deploying
WebPPL to generate a sample of 2,000 situation
descriptions. Table~\ref{tab:player_bat} shows the 
result. In the top row we see that with only selectional constraints,
the empirical probabilities of the two senses of
\textit{bat} are basically the same, as expected. The two lower rows
show the empirical probablities of \textsc{bat-animal} and
\textsc{bat-stick} when scenario constraints are present, showing a
clear preference for the \textsc{bat-stick} sense. This preference
grows more pronounced when the concentration parameter $\alpha$ of the
Dirichlet distribution is lower, that is, when we implement a stronger
preference towards sparse scenario distributions. Ideally, we would
tune the setting of $\alpha$ on human data to best match human
perceptions of word sense. Note that the
\textsc{bat-animal} sense is not ruled out with any setting of $\alpha$, it is just dispreferred,
more or less strongly.

\subsection{The astronomer sentence}

\begin{figure}[tb]
  \centering
  \begin{tabular}{c@{\hspace{4em}}c}
    \includegraphics[scale=0.4]{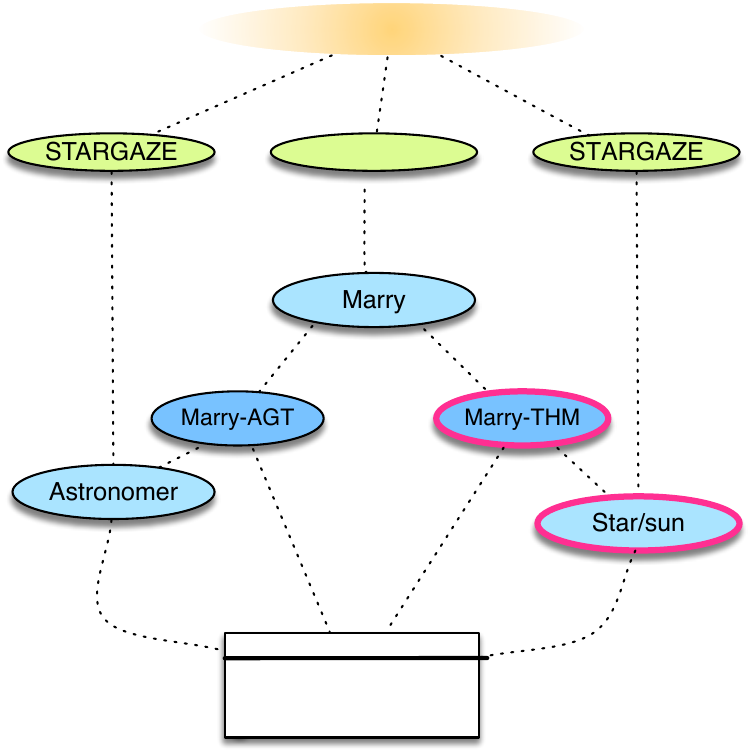}
    &
      \includegraphics[scale=0.4]{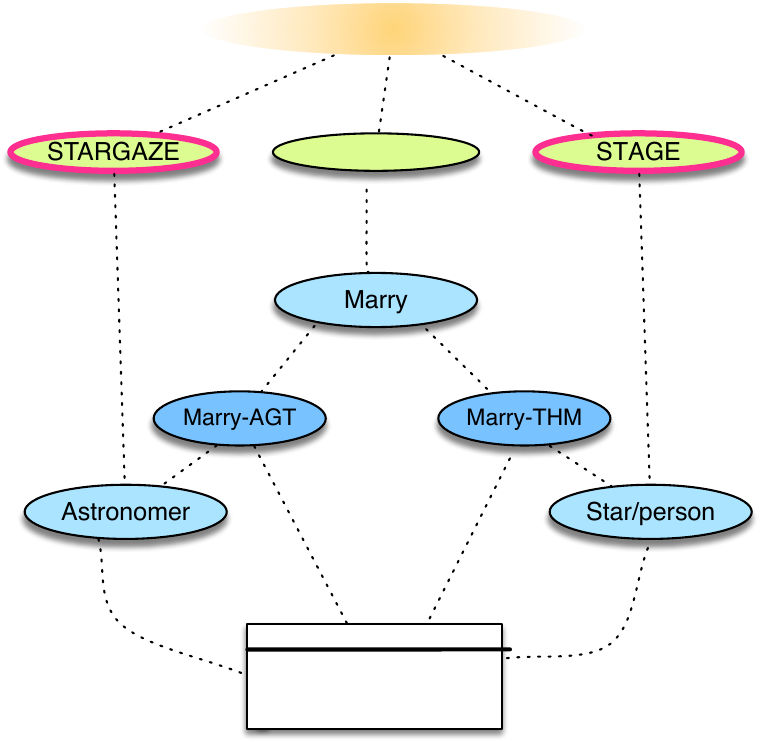}
  \end{tabular}
  \caption{Conflicting constraints in the sentence \textit{The astronomer married the star}: Either the concept for \textit{star} conflicts with the selectional constraint (left), or it conflicts with the preference for a coherent scenario (right)}
  \label{fig:astronomer_pic}
\end{figure}

\begin{table}[tb]
  \centering
  \begin{tabular}{l|l|l}
    $\alpha$ &\textsc{star-person}& \textsc{star-sun}\\\hline
    0.5& 0.82& 0.19\\
    0.1& 0.57 & 0.43\\
  \end{tabular}
  \caption{Conflicting constraints: Empirical probabilities for either a ``person'' or  a ``sun'' interpretation of \textit{star} in \textit{The astronomer married the star}, for different settings of the Dirichlet concentration parameter $\alpha$}
  \label{tab:astronomer}
\end{table}

The pun sentence from the beginning, repeated here
as \ref{ex:astronomer_illu_again}, has an SDS that is similar in structure
to that of \ref{ex:playerbat}. The point of this example is that
selectional preference and scenario constraints can conflict and
``pull in different directions,'' as illustrated in
Figure~\ref{fig:astronomer_pic}.

\ex. \label{ex:astronomer_illu_again} An astronomer married a star.

We again analyze this sentence experimentally through a WebPPL
simulation, with the following settings. We use two scenarios. The
scenario \textsc{stargazing} gives equal probabilities to the
concepts \textsc{astronomer}, \textsc{star(sun)}, and \textsc{marry},
and zero otherwise, 
while the scenario \textsc{stage} gives equal probabilities to
the concepts \textsc{star(person)} and \textsc{marry}, and zero otherwise (For
simplicity, we have added  \textsc{marry} to both scenarios instead of
adding a third scenario.)
The concept \textsc{marry} has mandatory Agent and Theme roles, both
with a strong preference for human role fillers: We set $P(c \mid
\text{\textsc{marry-theme}}) =0.475$ for a concept
$c=$\textsc{astronomer} or $c=$\textsc{star-person} and $P(c \mid
\text{\textsc{marry-theme}})=0.05$ for $c=$\textsc{star-sun}.
We again associate each concept with a single condition label, where the condition label is $star(\_)$ for
both \textsc{star-person} and \textsc{star-sun}.

Table~\ref{tab:astronomer} shows empirical probabilities for
\textit{star} being interpreted as either a person  or a sun, again
estimated from 2000 samples, for two different values for the
concentration parameter $\alpha$. Both $\alpha$ values generate a pun
effect, and the more emphasis there is on a coherent scenario (the
lower the value of $\alpha$), the more probability mass is given to
the situation where an astronomer marries a giant ball of plasma. 

\section{Probabilistic inferences about the world in Situation
  Description Systems}


The listener infers underlying concepts from the words in the
utterance. But conversely, the listener can also infer additional facts
about the situation from those concepts: If an entity is described as
an instance of a \textsc{bat-animal}, then it is definitely also an
animal, and there is a good chance that it is black and has fangs.

We
can formalize this idea as follows. So far, each concept-valued random
variable comes with a conditionally dependent variable for the
condition label used to name the concept. We can let the concept
sample additional DRS condition labels that do not name the concept
but provide additional properties of its instances. To do this, we
 add more nodes that
are conditionally dependent on the concept, one for each possible
property. The properties are probabilistic: all bats are animals,
but not all are black. The sampled DRS condition labels are then be
projected back into the DRS as additional conditions of the same
discourse referents.

We can also infer the presence in the situation of discourse
referents that are not mentioned in the utterance. The sentence
\textit{a girl was eating} does not mention what the girl was
eating. But if the role \textsc{eat-theme} is not observed, we might
still sample that it should be present in the sentence, and proceed to
sample a concept to fill it: Maybe what she was eating was a sandwich,
or an apple. Concepts describing such added discourse referents are
subject to all the constraints in the SDS, for example the imagined Theme of
\textsc{eat} needs to match the selectional preference, so it is most
likely a food item.

\section{Fine-grained sense distinctions}
\label{sec:argument}

\begin{table}[tb]
  \centering
\begin{tabular}{l|lllll}
    & 1:evidence & 2:quarrel & 3:pro/con & 5:parameter & 7:logical reasoning\\\hline
A1 & 3               & 5              & 4                     & 1  & 4\\
A2 & 1               &  5             & 2                      &1   & 2\\
A3 &1               & 2            &   3                    &1    & 4\\
  \end{tabular}
  \caption{Sentence \textit{She seems to revel in  arguments and loses no opportunity to
declare her political principles:} Ratings by three annotators (A1,
A2, A3) on how strongly each WordNet sense of \textit{argument}
applies. Ratings are on a scale of 1-5, where 5 is ``fits completely''
and 1 is ``does not fit at all''. See the text for more information
on the
senses.} 
  \label{tab:argumentsent}
\end{table}

So far, we have only considered examples that involved homonymy or at
least widely distinct senses. In this section, we consider an example with closely related
senses, the word \textit{argument} in sentence \ref{ex:argument_rep},
repeated from \S\ref{sec:generative}. 

\ex. \label{ex:argument_rep} She seems to revel in  \textit{arguments} and loses no opportunity to
declare her political principles.

In the WordNet dictionary~\citep{Fellbaum:98}, the noun
\textit{argument} has 7 senses including the following:\footnote{We
  are omitting senses 4 and 6, which are very specialized. Sense 4
  is a summary of a literary work. Sense 6 is from
  mathematics and is a variable
  on which the dependent variable depends.}
\begin{description}
\item[argument\#1] a fact or assertion offered as evidence that
  something is true: ``it was a strong argument that his hypothesis was true'' 
\item[argument\#2] a contentious speech act; a dispute where there is
  strong disagreement: ``they were involved in a violent argument'' 
\item[argument\#3] a discussion in which reasons are advanced for and
  against some proposition or proposal: ``the argument over foreign aid goes on and on'' 
\item[argument\#5] (computer science) a reference or value that is
  passed to a function, procedure, subroutine, command, or program 
\item[argument\#7] a course of reasoning aimed at demonstrating a truth
  or falsehood; the methodical process of logical reasoning: ``I can't
  follow your line of reasoning'' 
\end{description}

It is arguably genuinely unclear whether sentence \ref{ex:argument_rep}
says that the referent of the pronoun \textit{she} likes quarrels
(sense 2), debates (sense 3), or lines of reasoning (sense 7) -- or
some combination of these options. In \citet{erk-etal-2009-investigations},
we asked annotators to do word sense annotation with graded scores
over a dataset which included sentence \ref{ex:argument_rep}. Table~\ref{tab:argumentsent}
shows scores, on a scale of 1-5, that three annotators gave for the 
word \textit{argument} in that sentence. Annotator A1 rates all three senses
highly. A2 gives a high rating to the ``quarrel'' sense but not the
other two, and annotator A3 ranks the senses as first ``line of
reasoning,'' then ``debate,'' then ``quarrel.''\footnote{As a side note, there are
  several sentences in the dataset where the annotators disagree about
  sense 2 of \textit{argument}, ``quarrel'', where one annotator sees
  the sense as applying very often, but the other annotators do
  not. There is also repeated disagreement about sense 3, 
``debate.''}

\paragraph{An SDS analysis for the argument sentence.}

How can we explain the different annotators' intuitions? One possibility is
via selectional preference. A colleague commented
that in his opinion, one can only \textit{revel} in high-energy
activities, such as quarreling.  Such a preference could
explain the ratings of annotator A2, who gave a much higher score to
the ``quarrel'' sense of argument than to all others.

Another
possibility is that words such as \textit{political} and
\textit{principles} tend to conjure up a debate or opinion
article, in which case the ``pro-con'' and ``line of reasoning''
senses would be a better fit. This is one possible explanation for the
ratings of annotator A3. In an SDS, we would model this influence
through scenario constraints.

Another possible explanation is overall sense prevalence: The senses
of a word typically differ in overall frequency, and it stands to
reason that listeners should be sensitive to this fact. In SDSs, one
way to model sense frequency is through the frequency of its
underlying scenario: When a listener frequently encounters a scenario
$s$, then they also frequently encounter the concepts that are
prevalent in $s$.\footnote{Without further analysis we cannot know which of
these three reasons, if any, underlie the ratings that the annotators
gave. But it is possible, in principle, to investigate further: elicit
descriptions of selectional preferences along the lines of
\citet{McRaeThemrolesAsConcepts}, count corpus co-occurrences for
words like \textit{quarrel} or \textit{debate} with
\textit{political}, or obtain sense frequencies from annotated data.}

We again run some computational simulations to see if we can obtain
approximations of the ratings that the three annotators gave. To do
this, we first simplify the sentence so it will fit with our
current DRT fragment. We rewrite the sentence to
\ref{ex:argumentsimple}. In addition, we ignore the identity of discourse
referents. For example, the Experiencer of \textit{revel} is the Agent of
\textit{declare}, but we treat them as if they were separate
discourse referents. We also do not distinguish singulars
and plurals. Table~\ref{tab:argument_drs} shows the simplified DRS. 

\ex. \label{ex:argumentsimple} She reveled in arguments and always
declared her political principles.

\begin{table}[tb]
  \centering
\drs{$x_1$, $x_2$, $y$, $e_1$, $e_2$, $e_3$, $z_1$, $z_2$, $s$}
{female$(x_1)$,
  argument$(y)$,\\
  revel$(e_1)$, 
  Experiencer$(e_1, x_1)$,
  Stimulus$(e_1, y)$,\\
  principle$(z_1)$,  political$(s)$,
  Topic$(s, z_2)$,\\
  declare$(e_2)$,  Agent$(e_2, x_2)$,
  Theme$(e_2, z_1)$, always($e_3$)}
  \caption{Simplified DRS for sentence \ref{ex:argument_rep} which we use
    in our analysis}
  \label{tab:argument_drs}
\end{table}

We focus on senses 2, 3, and 7 of \textit{argument}, and represent
them as concepts, with an overall concept inventory of
\begin{quote}
\textsc{argument-quarrel}, \textsc{argument-debate},
\textsc{argument-logical},  \textsc{woman}, \textsc{principle},
\textsc{political}, \textsc{always}, \textsc{revel}, \textsc{declare}
\end{quote}
We add three matching scenarios
\textsc{sc-quarrel}, \textsc{sc-debate}, and \textsc{sc-logical}.
Because of the larger graphical model and the larger inventories of
concepts and scenarios, we switch from rejection sampling to
sequential Monte Carlo with 15,000 samples. 

\begin{table}[tb]
  \centering
  \begin{tabular}{l|ccc}
    Setting & quarrel & debate & logical\\\hline
    No preference & 0.32 & 0.35 & 0.33\\
    Scenario preference towards ``debate'' and ``logical'' & 0.13 & 0.41 & 0.46\\
    Selectional preference of \textit{revel} for ``quarrel'' & 0.74 & 0.12 & 0.14\\
    Scenario + selectional & 0.52 & 0.23 & 0.25\\
    Sense frequency  & 0.42 & 0.35 & 0.23\\
  \end{tabular}
  \caption{Computational simulation results for the \textit{she revels
      in arguments} sentence: probabilities for the senses
    \textsc{argument-quarrel}, \textsc{argument-debate} and
    \textsc{argument-logical} in different SDS settings}
  \label{tab:argument_results}
\end{table}

Table~\ref{tab:argument_results} shows results for different
settings. When \textit{revel} has equal preference for all senses of
\textit{argument}, and all three scenarios can sample all of the
concepts equally well, we obtain equal probabilities for all senses,
shown in the ``no preference'' row. If we assume that
\textit{political} and \textit{principle} fit the scenarios
\textsc{sc-debate} and \textsc{sc-logical}, but not
\textsc{sc-quarrel}, we obtain the results in the row labeled
``scenario preference'', where the ``debate'' and ``logical'' senses of
\textit{argument} are about equally strong, and much stronger than the
``quarrel'' sense. If we follow the intuition that \textit{revel}
favors high-energy activities such as quarrelling, we get the
results in the ``selectional preference'' row, with a probability of
0.74 for the ``quarrel'' sense of argument. (For this row,
\textsc{argument-quarrel} has been given a preference of 0.7, with
values of 0.1 for the other two senses of \textit{argument} and
\textit{principle}.) Combining the preference for the
\textsc{sc-debate} and \textsc{sc-logical} scenarios with the
selectional preference for \textsc{argument-quarrel}, we arrive at the
row labeled ``scenario+ selectional'', where the probability of the ``quarrel''
sense, at 0.,52, is still the highest but no longer as starkly
different from the others. Finally, the ``sense frequency'' row shows the
result of assuming asymmetric prior probabilities for the scenarios~\citep{NIPS2009_0d0871f0},
which translate into prior probabilities for the senses of
\textit{argument}. (For this row, we set the priors at 0.4 for
\textsc{sc-quarrel}, 0.35 for \textsc{sc-debate} and 0.25 for
\textsc{sc-logical}. We follow the sense order in WordNet, which was
set by counting sense frequencies in a sense-annotated corpus.) Of all
these results, the ``no preference'' experiment most closely resembles
the ratings of annotator A1, the
``selectional preference,'' ``scenario +  selectional'' and ``sense frequency'' experiments are the best
match for A2, and ``scenario preference'' is closest to the ratings of
A3.\footnote{One thing to note about our current system  is that it can only produce SDs that each select
a single sense. So, we cannot distinguish between a listener who
thinks several senses apply at the same time (an AND of senses), and a listener who
cannot decide between two different and exclusive readings of the
sentence (an OR of senses). One could envision an extension of SDS
where a word token could be associated with a group of concepts; but
it is not clear that it is useful to have such a model, as we have previously found that it is
hard for annotators, in word sense annotation, to distinguish between
the AND and the OR case.}

\section{Extensions}
\label{sec:extensions}

The example sentence from Ex.~\ref{ex:argument_rep} above contains a variety of phenomena, including coreference, modification, genericity, habituality, plurality and negation. 
As it stands, our framework only offers a fragmentary conceptual representation of a given logical form. Full integration with DRT would require an extended framework to deal with core aspects of the logic. We cover three such aspects below, and give hints as to how to handle them.

First, we will require additional representations  so that 
important phenomena such as quantification or genericity can be
modelled at a conceptual level. This brings  in further questions, in
particular whether we want situation descriptions to include full
mental representations of instances, events, and groups thereof. If
so, the framework would have to be considerably extended to allow for
groups of individuals to be generated from a concept, each group with
their specific properties. But there is in principle no fundamental
issue in adding a probabilistic layer between the level of concepts
and the level of observed lexical items, linking sampled individuals
to denotations in the DRS.


Second, once we have representations of referents, the system should support dynamic updating of those representations. Updates would allow us to model changes in the speaker's knowledge about an individual brought in by each new piece of discourse. We note that modification could be tackled in the same manner, by first generating a relevant situation and concept for the referent, and updating the initial representation in light of the modifier.


Third, we note that the framework provides representations that are sufficiently complex to encode different types of negation. One interesting aspect of negated propositions is that they bring a certain conceptual content to the mind of the listener before making it logically false, e.g. \textit{a bat does not sleep} clearly brings to mind a sleeping bat, before stating that there is no such individual in the universe under consideration. This can naturally be modelled by having a situation description with a sleeping bat and assigning it a $0$ probability, whilst also taking care of generating alternative SDs with awake bats and non-zero probabilities. 

Ideally, SDSs would be set up so they can be used for evaluating the
consequences of certain choices in the formal semantics literature at
the cognitive level. For instance, we might want to implement
different proposals for the formalisation of genericity and observe
how they affect the production of conceptual content.


\section{Different frameworks for a fine-grained representation of lexical meaning, and where SDSs fit in}

Our Situation Description Systems are part of a more general effort to figure out how to integrate fine-grained representations of lexical meaning into sentence meaning representations. There is no agreement on a single framework at this point, not even an agreement on a general direction. This makes it hard to sum up and categorize all the diverging strands. One of us has tried several times, differently every time~\citep{Erk:Nefdt,erkAnnurev}. For the purpose of this paper, we will attempt a classification based on the main phenomena being addressed. 

Several frameworks have been proposed to describe fine-grained lexical meaning differences, using types~\citep{asher2011}, attribute-value matrices~\citep{Zeevat:RepresentingTheLexicon}, qualia~\citep{delpinalMeaningModulationContext2018} and, frequently, distributional models automatically computed from corpus data~\citep{grefenstette-sadrzadeh:2011:EMNLP,Baroni:FregeInSpace,Erk:alligators,AsherEtAl:CL,McNallyBoleda,emerson-2020-autoencoding,herbelot-2020-solve}. 

At the lexical level, one main question is how to model the variability of meaning in context: polysemy~\citep{asher2011,Zeevat:RepresentingTheLexicon} and vagueness~\citep{Sutton:Vague}. This also involves  interactions of different constraints that affect word meaning~\citep{Emerson,emerson-2020-autoencoding} and selectional preferences~\citep{chersoni_santus_pannitto_lenci_blache_huang_2019}. Another main question is about learning: How can lexical meaning be learned~\citep{larsson:jlc}, and what properties can be acquired~\citep{herbelot-2020-solve,HerbelotCopestake}?

Going beyond individual words, the issue of compositionality is maybe the one that has seen most work in this area: the combination of feature representations of smaller phrases into feature representations of larger phrases~\citep{grefenstette-sadrzadeh:2011:EMNLP,Baroni:FregeInSpace,paperno-etal-2014-practical,SadrzadehMuskens:18}. But another important and difficult question concerns the interaction of feature-based lexical representations with quantifiers and negation \citep{Bernardy:18,bernardy-etal-2019-bayesian,bernardy-etal-2019-predicates,EmersonQuantified}.
The exploration of fine-grained lexical representations has also raised again questions related to the nature of sentence meaning representations. Several research efforts use a combination of intensional and conceptual representation~\citep{asher2011,GoodmanLassiter,McNally:2017tr,Pelletier:2017,delpinalMeaningModulationContext2018,Emerson}. 

Situations are another recurring theme in several approaches, in particular the importance of (probabilistically) imagined situation in sentence understanding~\citep{Sutton:Vague,GoodmanLassiter,Cooper:2015vj,venhuizen2021distributional}, 

With these main questions in mind, we can now situate Situation Description Systems with respect to these other approaches. We formulate SDSs as a dual meaning representation that is both intensional and conceptual. SDSs focuses on different and interacting constraints on meaning in context, like \citet{chersoni_santus_pannitto_lenci_blache_huang_2019} and \citet{emerson-2020-autoencoding}. We suggest that it is important to include scenario knowledge among those constraints. This links SDSs to approaches that foreground situations, though SDSs differ from the other approaches in viewing scenarios as related to the frames in Fillmore's Semantics of Understanding ~\citep{Fillmore:Usemantics}.

\section{Conclusion}


In this paper, we have proposed a framework for describing word meaning in context. Our account regards  sentence understanding as a process that integrates the concepts and scenarios evoked by the words in the sentence, guided by local and global constraints. 
 We have argued that this integration mechanism naturally accounts for the specialization of context-independent lexical meanings into token meanings. Our examples throughout the paper highlight particular characteristics of meaning in context, which we briefly summarize next.

 First, meaning in context calls on phenomena beyond word senses: in particular, the underlying scenarios play an important role in interpretation. Second, token meaning does not end at the token but involves a network of constraints through which meanings are inextricably linked to each other. Third, it is not enough to identify a single prevalent sense for each word in the sentence. In some cases, there are several senses that might fit a token, and interpretation involves the awareness of \textit{all} of them.

Going forward, there are many ways in which our formalisation will need to be extended. This includes the issues of dynamic interpretation, quantifiers, and negation discussed above. In addition, we will need to develop an account of semantic construction for our framework that describes how \textit{both} the DRS and the graphical model for a sentence can  be assembled incrementally. Finally, as previously mentioned, we must work out a scalable implementation of the framework that will retain our ability to trace the interpretation of a given utterance in an explainable manner.

We hope, at any rate, that our proposal provides a stepping stone for developing a description of what Fillmore had in mind when he talked about ``interpretation of the whole'' \citet[p. 233]{Fillmore:Usemantics} in the process of utterance understanding. 
As we have seen, formalizing the process has 
consequences for the way we define meaning, in the lexicon and beyond.
We believe that attempting a formal description of comprehension can fruitfully contribute to their elucidation.


\bibliography{astronomer1} 

\pagebreak
\appendix
\section{Appendix: Overview of the sampling process}

A directed graphical model can be characterized by a ``generative
story''. This is a summary of the sampling process from the graphical
model, following directed edges from the top down. For the situation
description systems that we have used in this paper, the generative
story runs as follows. We use greek letters for parameters, and
sequences of parameters, for probability distributions. We use
subscripts $\cdot_s$, $\cdot_c$, $\cdot_{c, r}$ to indicate that a parameter is
specific to a scenario $s$, concept $c$, or role $r$ of concept $c$.

\begin{itemize}
\item Draw a distribution $\theta$ over scenarios from a Dirichlet
  distribution with concentration parameter $\alpha$. 
\item From a discrete uniform distribution over numbers $1, \ldots,
  N$, draw $n$, the size of the situation description. This is the
  number of top-level concepts we will sample.
\item Do $n$ times:
  \begin{itemize}
  \item Sample a scenario $s$ from a multinomial distribution with
    parameter $theta$.
  \item Sample a concept $c$ from a multinomial distribution with
    parameter $\phi_s$,
  \item For each role label $r$:
    \begin{itemize}
    \item Sample whether role $r$ is present for concept $c$ in the
      situation or not. Use a Bernoulli distribution with parameter
      $\rho_{c, r}$. If the answer is
      yes:
    \item Sample a scenario $s'$ from a multinomial distribution
      with parameter $\theta$
    \item Sample a role filler concept $c'$ jointly from a multinomial
      distribution with parameter $\phi_s$ (the concept distribution
      of scenario $s$) and from a multinomial distribution with
      parameter $\phi_{c, r}$ (the selectional constraint)
    \end{itemize}
  \end{itemize}
\item For each concept token $c$: Sample a unary DRS condition label
  describing the concept, from a multinomial distribution over
  DRS condition labels with  parameter $\xi_c$ 
\item For each role token $\langle c, r\rangle$: Sample a binary DRS
  condition label describing the role label, from a multinomial
  distribution over DRS condition labels with a parameter $\xi_{c,  r}$. 
\end{itemize}

\end{document}